\documentclass[letterpaper]{article} 
\usepackage{aaai25}  
\usepackage{times}  
\usepackage{helvet}  
\usepackage{courier}  
\usepackage[hyphens]{url}  
\usepackage{graphicx} 
\usepackage{natbib}  
\usepackage{caption} 
\usepackage{algorithm}
\usepackage{algorithmic}
\usepackage{multirow}
\usepackage{amsmath}
\usepackage{amssymb}
\usepackage{enumitem}
\usepackage{newfloat}
\usepackage{listings}
\DeclareCaptionStyle{ruled}{labelfont=normalfont,labelsep=colon,strut=off} 
\floatstyle{ruled}
\newfloat{listing}{tb}{lst}{}
\floatname{listing}{Listing}
\title{Skeleton-based Action Recognition with Non-linear Dependency Modeling and Hilbert-Schmidt Independence Criterion}

\author{
    Haipeng Chen,\textsuperscript{\rm 1,2},
    Yuheng Yang\textsuperscript{\rm 1,2}\thanks{Corresponding to Yuheng Yang and Yingda Lyu.},
    Yingda Lyu\textsuperscript{\rm 1,3*}
}
\affiliations{
    \textsuperscript{\rm 1}College of Computer Science and Technology, Jilin University,\\
    \textsuperscript{\rm 2}Key Laboratory of Symbolic Computation and Knowledge Engineering of Ministry of Education, Jilin University\\
    \textsuperscript{\rm 3}Public Computer Education and Research Center, Jilin University\\
    chenhp@jlu.edu.cn, yangyh20@mails.jlu.edu.cn, ydlv@jlu.edu.cn 

}

\usepackage{bibentry}
\begin{document}

\maketitle

\begin{abstract}
Human skeleton-based action recognition has long been an indispensable aspect of artificial intelligence. Current state-of-the-art methods tend to consider only the dependencies between connected skeletal joints, limiting their ability to capture non-linear dependencies between physically distant joints. Moreover, most existing approaches distinguish action classes by estimating the probability density of motion representations, yet the high-dimensional nature of human motions invokes inherent difficulties in accomplishing such measurements. In this paper, we seek to tackle these challenges from two directions: (1) We propose a novel dependency refinement approach that explicitly models dependencies between any pair of joints, effectively transcending the limitations imposed by joint distance. (2) We further propose a framework that utilizes the Hilbert-Schmidt Independence Criterion to differentiate action classes without being affected by data dimensionality, and mathematically derive learning objectives guaranteeing precise recognition. Empirically, our approach sets the state-of-the-art performance on NTU RGB+D, NTU RGB+D 120, and Northwestern-UCLA datasets. 
\end{abstract}

%

\section{Introduction}

\label{sec:intro}
Endowing machines with the ability to perceive and recognize human behaviors is very much coveted for various applications ranging from virtual reality to security monitoring~\cite{liu2022copy,wu2024pose}. Consequently, action recognition has attracted much interest, particularly for methods that rely on skeletal data, which is robust against environmental noise and viewpoint changes.

Recent skeleton-based approaches ~\cite{yan2018spatial,shi2019skeleton,shi2019two} tend to employ Graph Convolutional Networks (GCNs) to model human motion patterns since the hierarchical and tree-like graph structure naturally in the human skeleton. For instance, 
~\cite {xu2022topology} attempts to design sophisticated adjacency matrices, seeking to pursue more nuanced modeling of spatial joint dependencies.
~\cite{chi2022infogcn,zhou2023learning} put their efforts into learning discriminative motion features from the skeleton sequence. ~\cite{yang2023action} adopts an information-theoretic objective to fully mine task-relevant information while reducing task-irrelevant nuisances. 

\begin{figure}[t]
  \centering
\includegraphics[width=\linewidth]{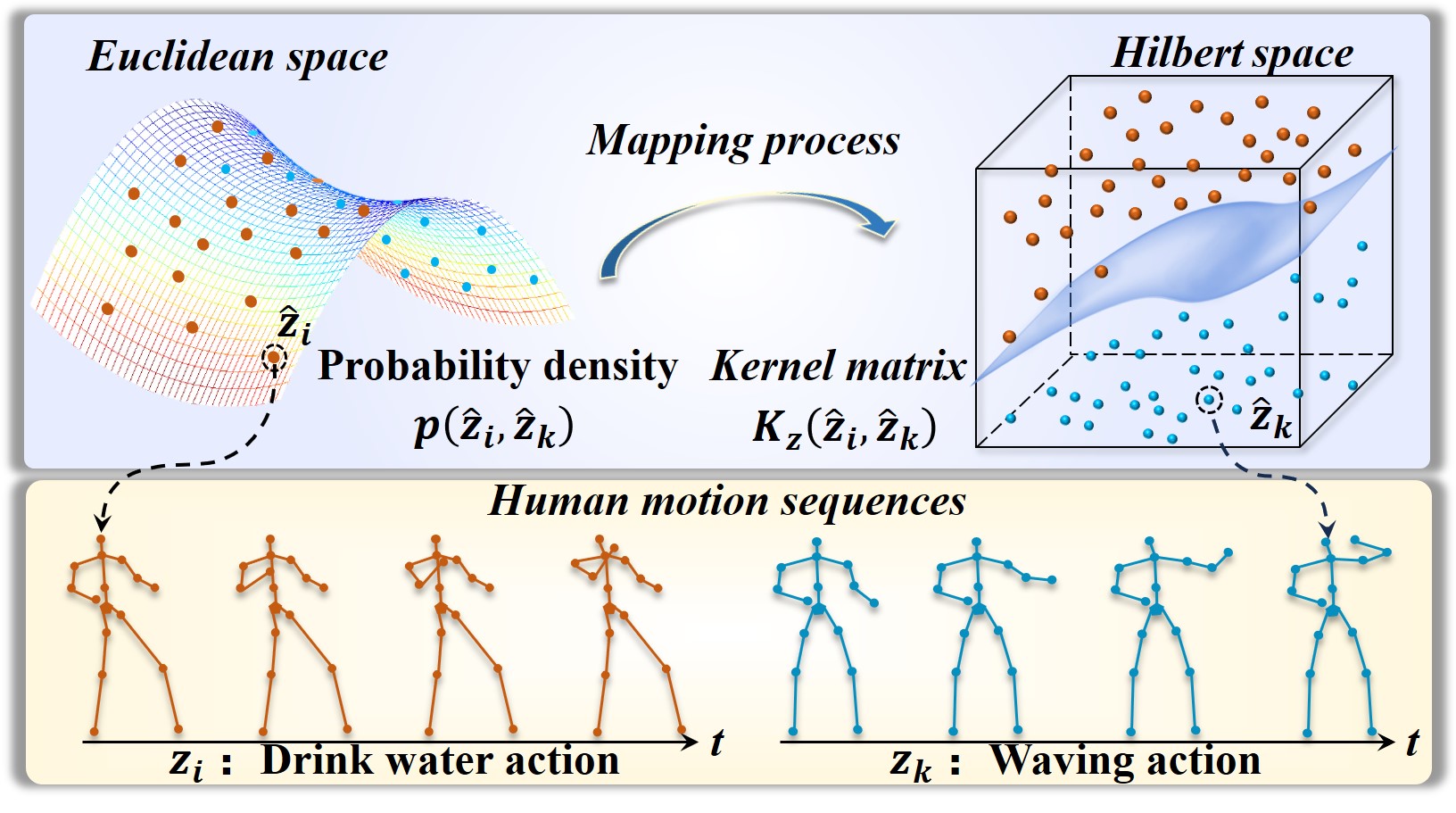}
  \caption{The conceptual diagram illustrates the mapping process of feature representations from Euclidean space into Hilbert space. The mapping process is achieved through the kernel function. }
  \label{fig1}
\end{figure}
Unfortunately, after a systematic investigation of prior works, we observe that they still suffer from inaccurate action recognition. We conjecture that the reasons are two folds.
\textbf{(1) First}, current methods~\cite{yan2018spatial,li2019actional} typically model human motions by performing graph convolutions on pre-defined skeleton graphs. However, these approaches exhibit a severe limitation as they consider only the dependency between physically connected joints while ignoring non-linear dependencies between geometrically separated joints. Although several works~\cite{xu2022topology,song2022constructing} tend to address this through hierarchical graphs or scaling graphs, they still struggle to effectively model non-linear dependencies between distant joints due to the limited receptive field of graph convolutions.
\textbf{(2) Second}, 
given the motion feature representation, existing approaches~\cite{chi2022infogcn,yang2023action,zhou2023learning,huang2023graph} typically estimate its probability density in conjunction with the category label to identify the action class. A higher probability density indicates a stronger likelihood of the motion feature belonging to the correct class. Nevertheless, a motion sequence is characterized as a time series of skeletal poses, each of which translates to the positions of all joints, resulting in a \emph{high-dimensional} representation. Estimating the probability density of these \emph{high-dimensional} representations in the raw Euclidean space introduces unnecessary training difficulty, which leads to reduced accuracy in action classification.

In this paper, we embrace two key components to tackle these challenges. Technically, \textbf{(i)} We propose a dependency refinement method that strips the skeletal structure down to dynamically zoom in on the \emph{joint-to-joint} dependencies of pairwise skeletal joints. Specifically, setting aside the skeletal structure, we explicitly model each pair of joint dependencies with a Gaussian correlation function. 
By adjusting the kernel width in the Gaussian correlation function, we could fine-grainedly control the influence of distance on joint dependencies. These dependencies are then used to adaptively refine the initial skeletal graph, enabling precise human motion modeling. We also employ an ensemble of networks trained with different kernel widths, seeking to improve the comprehensiveness and accuracy of action recognition.
\textbf{(ii)} Based on the aforementioned method, we further propose a novel framework that leverages the Hilbert-Schmidt Independence Criterion (HSIC) for facilitating action recognition.  First, we utilize a Hilbert kernel function to map high-dimensional motion features from the straightforward Euclidean space to a Hilbert space, as illustrated in Fig.~\ref{fig1}. In this space, the HSIC mathematically evaluates the statistical dependence between these features and the corresponding action labels. 
Second, we theoretically derive the learning objectives, guaranteeing the efficacy of the final action classification. We would like to point out that, since HSIC is defined in terms of kernel method, the process of distinguishing action classes operates in a \emph{dimension-independent} manner.

Thereafter, we conduct extensive experiments on three popular benchmark datasets, namely the NTU RGB+D 60 dataset, the NTU RGB+D 120 dataset, and the Northwestern-UCLA dataset. 
The empirical results show that our approach consistently and significantly outperforms state-of-the-art performance.
To summarize, our key contributions are as follows: 
\begin{itemize}
\item 
A dependency refinement method is presented to comprehensively learn the relationships between joints, which simultaneously considers the skeletal connection between adjacent joints and the non-linear dependencies between distant joints.



\item 
We propose a novel action recognition framework, which effectively distinguishes action classes among high-dimensional motion feature representations while deriving learning objectives to ensure the efficacy of the action classification.


\item Our method outperforms existing approaches and achieves state-of-the-art performance on three popular benchmark datasets. 
The implementations have been released, hoping to facilitate future research.
\end{itemize}
\section{Related Work}
\textbf{Skeleton-based action recognition.} 
Previous methods tackle the skeleton-based action recognition task by utilizing Convolutional Neural Networks (CNNs) or Recurrent Neural Networks (RNNs)~\cite{liu2016spatio,liu2019towards},  which unfortunately ignore the inherent relationships between joints. Recently, there has been an increasing interest in developing Graph Convolutional Networks (GCNs)~\cite{yan2018spatial} since the advantages in dealing with irregular graphical structures. 
FR-Head~\cite{zhou2023learning} introduces an auxiliary feature refinement head to acquire discriminative representations of skeletons, aiding in distinguishing ambiguous actions. 
Stream-GCN~\cite{yang2023action} engages the mutual information loss to maximize the task-relevant information while minimizing the task-irrelevant nuisances for facilitating the action recognition. SkeletonGCL~\cite{huang2023graph} introduces graph contrast learning to explore the cross-sequence global context in a fully supervised setting.

\noindent\textbf{Hilbert-Schmidt independence criterion.}
The Hilbert-Schmidt Independence Criterion (HSIC) is a statistical measure used to assess the independence between two random variables~\cite{gretton2005measuring}. HSIC maps input data into feature vectors and computes their product to evaluate correlation. Its strength lies in uncovering intricate data correlations within a Reproducing Kernel Hilbert Space (RKHS) without being constrained by the dimensionality of the input data~\cite{bertsimas2022data}.  
For instance, HSIC-InfoGAN~\cite{liu2022hsic} learns unsupervised disentangled representations by directly optimizing the Hilbert-Schmidt Independence Criterion (HSIC) loss, eliminating the need for an auxiliary network. 
To the best of our knowledge, there is no related research on the utilization of HSIC in the skeleton-based action recognition task.
\section{Methodology}
\textbf{Notations.}\quad Generally, the human skeleton can be regarded as an interlinked structure comprised of joints and bones. We denote the set of joints as a set of nodes $\mathcal{V}=\{v_1, \dots, v_N\}$. Additionally, we represent the set of bones as a set of edges $\mathcal{E}$, which can be formulated as an adjacency matrix $\textbf{A} \in \mathbb{R}^{N \times N}$. Hence, we conveniently depict a human skeleton as a graph $\mathcal{G}=(\mathcal{V}, \mathcal{E}$). An action can be represented as a sequence of skeleton poses  $\mathcal{X} = \langle x_1, x_2, \cdots, x_T \rangle$, where $x_t$ denotes the skeleton pose at time $t$ and $T$ is the number of frames. Since $x_t \in \mathbb{R}^{N \times C}$, where $N$ is the number of joints, $C$ denotes the dimension of a joint, the 3D motion sequence $\mathcal{X}$ can be formulated as a feature tensor $\textbf{X} \in \mathbb{R}^{T \times N \times C}$. Presented with a 3D skeleton motion sequence $\mathcal{X}$, we aim to accurately predict its action class label $y$.


%

\subsection{Non-linear Dependency Modeling}\label{3.3}
\begin{figure}[t]
  \centering
  \includegraphics[width=\linewidth]{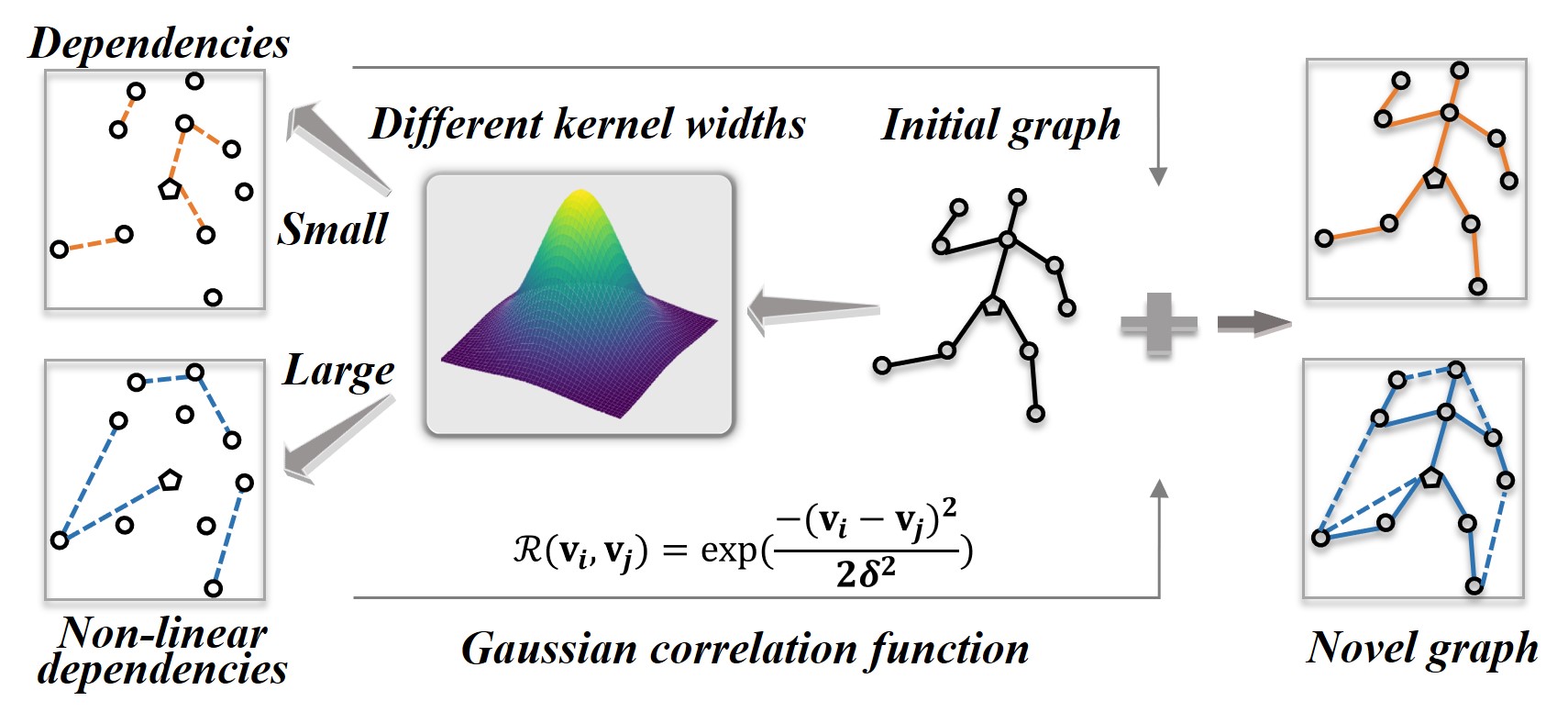}
  \caption{The diagram illustrates the dependency refinement method. Specifically, we utilize the Gaussian correlation function to quantify dependencies between joints and incorporate them into the initial graph. By adjusting the kernel width in the Gaussian function, we could effectively capture the dependencies at both adjacent (orange) and distant (blue) scales.}
  \label{fig:parts} 
\end{figure}
Current works~\cite{shi2019two,liu2020disentangling,cheng2020decoupling} typically learn human motions by modeling the dependencies between adjacent joints. Unfortunately, these methods ignore the relationships between geometrically separated joints. While some approaches introduce adaptive graphs or hierarchical graphs~\cite{lee2023hierarchically,wu2024joint} to mitigate such problems, they are known to have difficulties in capturing the non-linear dependencies between distant joints due to the limited receptive field of graph convolutions. 
Motivated by these insights, we propose a novel dependency refinement method designed to overcome the limitations imposed by joint distance, allowing for precise human motion modeling. In what follows, we will delve into the specifics of our approach.

Technically, for the implementation of graph convolution, we define inputs $\mathcal{X}$ represented by features $\textbf{X}$ and graph structure $\textbf{A}$. The update rule of GCNs is given by:
\begin{equation}
\textbf{X}^{(s)} = \sigma\left(\textbf{D}^{-\frac{1}{2}}(\textbf{A}+\textbf{I})\textbf{D}^{-\frac{1}{2}}\textbf{X}\textbf{W}\right),
\end{equation}
where $\sigma$ indicates an activation function like ReLU, $\textbf{X}^{(s)} \in \mathbb{R}^{T \times N \times C'}$ denotes the extracted spatial features, $\textbf{D}$ is the diagonal degree matrix of $\textbf{A}+\textbf{I}$, and $\textbf{W}$ represents the learnable weights for feature transformation. 
Subsequently, we leverage the Gaussian correlation function $\mathcal{R}(\cdot)$ to model joint dependencies. As shown in Fig.~\ref{fig:parts}, given a pair of joints $(v_i, v_j)$, the features of ${v}_i$ and ${v}_j$ can be formulated as $(\textbf{v}_i, \textbf{v}_j)$, where $\textbf{v}_i$, $\textbf{v}_j \in \mathbb{R}^{C}$, and $i$, $j$ indicate the indices of joints. We then input the features $\textbf{v}_i$ and $\textbf{v}_j$ into $\mathcal{R}(\cdot)$, formulated as: 
\begin{equation}
\begin{aligned}
\mathcal{R}(\textbf{v}_i, \textbf{v}_j) = \exp\left(\frac{-(\textbf{v}_i-\textbf{v}_j)^2}{2\delta^2}\right),
\end{aligned}
\end{equation}
where $\delta$ represents the kernel width. Based on the Gaussian correlation function, we obtain the dependencies with a linear transformation $\phi$:
\begin{equation}
\begin{aligned}
\textbf{r}_{ij} = \phi \left(\mathcal{R}(\textbf{v}_i, \textbf{v}_j)\right), i,j \in \{1, 2, \cdots, N\},\label{eq:gau}
\end{aligned}
\end{equation}

where $\textbf{r}_{ij} \in \mathbb{R}^{C'}$ reflects the dependency between ${v}_i$ and ${v}_j$. \textbf{Interestingly}, by adjusting the size of $\delta$, we can directly control the degree to which distance affects the dependency. For instance, when performing a complex action such as the ``Kicking" action, which requires coordination between the feet, legs, and torso, there are likely to be distant connections between these joints. In such cases, it is necessary to use a large $\delta$ so that the dependency decays more slowly as the distance between joints increases. This ensures that the influence of distance on the dependency is weakened, aiming joints that are structurally apart to still exhibit strong dependencies. 

Furthermore, we incorporate the dependencies from Equation~(\ref{eq:gau}) to refine the initial graph, yielding a novel GCN implementation:
\begin{equation}
\textbf{X}^{(s)}_c = \sigma\left(\textbf{D}^{-\frac{1}{2}}(\textbf{A}_c+\textbf{I})\textbf{D}^{-\frac{1}{2}}\textbf{X}\textbf{W}\right),
\end{equation}
where $\textbf{A}_c=\textbf{A}+ \textbf{W}_c\textbf{R}$, $\textbf{W}_c$ denotes the learnable parameters for the dependency, $\textbf{R} \in \mathbb{R}^{N \times N \times C'}$ denotes the dependency matrix consists of $\textbf{r}_{ij}$, and $\textbf{A}_c$ is the novel adjacency matrix obtained in a broadcast manner. 

\begin{figure*}[t]
\centering
\includegraphics[width=\linewidth]{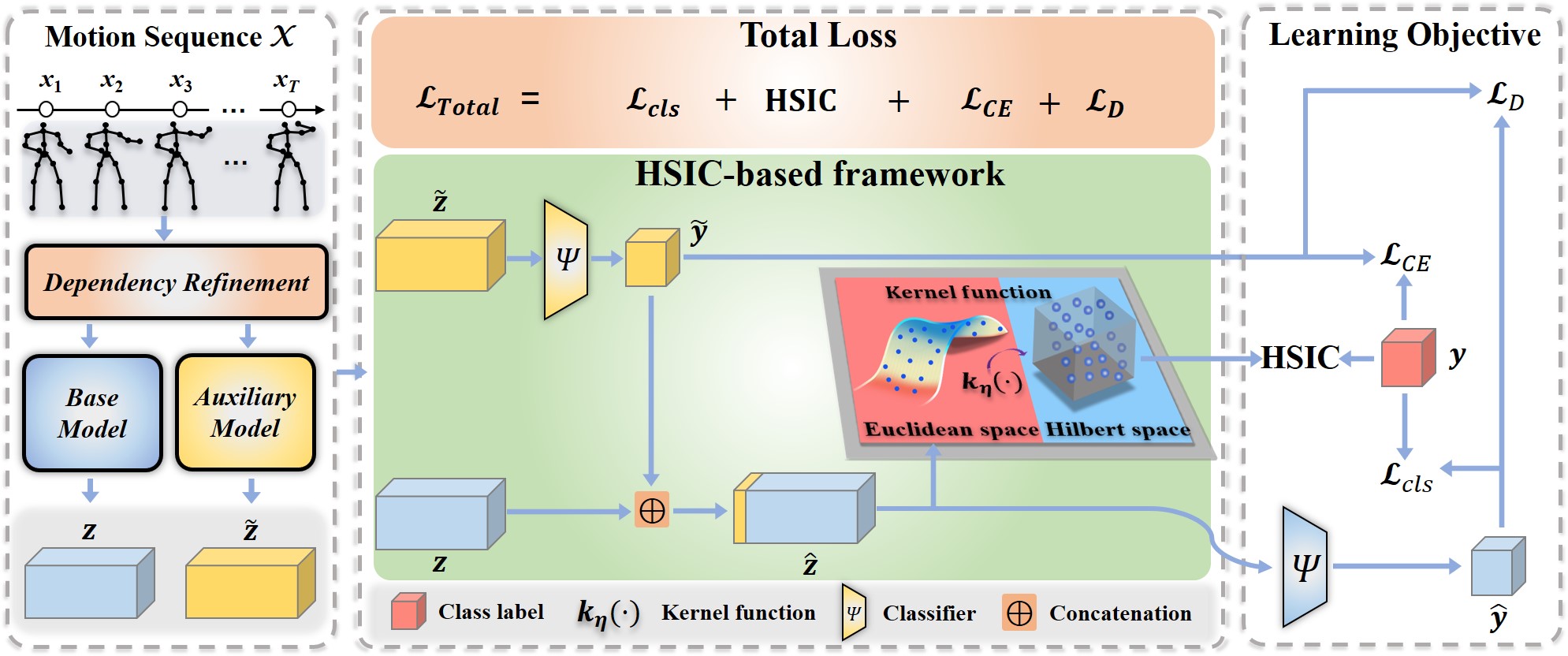}
\caption{The overall pipeline of our HSIC-based framework aims at recognizing the action classes of the motion sequences. For clarity, we only illustrate the pipeline using a single sequence  $\mathcal{S}$ in this figure. The pipeline starts with refining joint dependencies, followed by extracting the motion features $z$ and $\tilde{z}$ from the base model and the auxiliary model, respectively. Subsequently, we feed $\tilde{z}$ to a classifier to obtain the auxiliary information $\tilde{y}$. In order to enhance the discriminative power of $z$, we incorporate $\tilde{y}$ into $z$ to obtain the augmented feature $\hat{z}$. We then engage a kernel function $k_\eta(\cdot)$ to transform $\hat{z}$ into Hilbert space and derive learning objectives, which effectively avoid the issue arising from the data dimensionality. The entire learning objective $\mathcal{L}_{Total}$ consists of $\mathcal{L}_{cls}$, HSIC, $\mathcal{L}_{CE}$, and $\mathcal{L}_{D}$.}
\label{fig:framework}
\end{figure*}
\subsection{Distinguishing Action Classes For Motion Feature Representations}
\label{3.2} 
Many existing methods~\cite{yang2023action,huang2023graph,zhou2023learning} advocate for distinguishing action classes by estimating probability density among multiple motion feature representations. However, factors in the motion sequence, such as \emph{time series}, \emph{3D joint coordinates}, and \emph{the number of joints}, result in a \emph{high-dimensional} feature representation. In such a high-dimensional space, feature representations tend to be sparsely distributed, leaving much of the sampling space unfilled, which makes it difficult to accurately classify the action classes~\cite{majdara2022efficient}.


To tackle this problem, we propose a framework based on the Hilbert-Schmidt Independence Criterion. As shown in Fig.~\ref{fig:framework}, our action recognition framework includes a base model for generating motion features and an auxiliary model for producing auxiliary motion information. By incorporating the auxiliary motion information into the motion features, we could explicitly enhance their discriminative ability. These enhanced features are then mapped from Euclidean space to Hilbert space using a Hilbert Reproducing Kernel Function, transforming them into kernel matrices.
Finally, we utilize HSIC values derived from these kernel matrices to identify action classes and mathematically formulate training objectives, ensuring effective learning without being hindered by high dimensionality.

Formally, our approach \textbf{initiates} by performing joint dependency refinement. \textbf{Then}, the base model is trained to encode the motion sequence $\mathcal{X}$ into the motion feature ${z}$. \textbf{Meanwhile}, 
the auxiliary model encodes $\mathcal{X}$ into $\tilde{z}$, which is then fed into the classifier to predict motion information $\tilde{y}$.
\textbf{Thereafter}, we incorporate $\tilde{y}$ into ${z}$ to obtain the enhanced feature $\hat{z}$ and utilize the Mat\'{e}rn covariance function as the \emph{Hilbert Reproducing Kernel Function} to map $\hat{z}$ into Hilbert Space. The kernel function is:
\begin{equation}
k_\eta(\hat{z}_u, \hat{z}_w) = \alpha \frac{2^{1-\eta}}{\Gamma(\eta)} \left(\frac{\sqrt{2\eta} r}{\ell}\right)^\eta F_\eta \left(\frac{\sqrt{2\eta} r}{\ell}\right),
\end{equation}
where $\eta$ $>$ 0 is the kernel order, $F_\eta$ is the modified Bessel function of the second kind, $\alpha$ $>$ 0 and $\ell$ $>$ 0 denote the amplitude and length scale, $\Gamma(\eta)$ is the normalization factor, and $r=\|\hat{z}_u - \hat{z}_w\|_{2}$. $\{(\hat{z}_u, \hat{z}_w)\}_{u,w=1}^n$ denote $u^{th}$ and $w^{th}$ feature samples. The particular case where $\eta=\frac{3}{2}$ is probably the most commomly used kernel~\cite{williams2006gaussian}: 
\begin{equation}
k_\frac{3}{2}(\hat{z}_u, \hat{z}_w) = \alpha \left(1+ \frac{\sqrt{3}r}{\ell}\right)\exp \left( -\frac{\sqrt{3}r}{\ell} \right).
\end{equation}

\textbf{Subsequently}, we could obtain the kernel matrix $K_{\hat{z}}$, which is defined as follows:
\begin{align}
K_{\hat{z}} &= \begin{pmatrix}
k_\frac{3}{2}(\hat{z}_1, \hat{z}_1) & \dots & k_\frac{3}{2}(\hat{z}_1, \hat{z}_n) \\
k_\frac{3}{2}(\hat{z}_2, \hat{z}_1) &  \dots & k_\frac{3}{2}(\hat{z}_2, \hat{z}_n) \\
\vdots & \ddots & \vdots \\
k_\frac{3}{2}(\hat{z}_n, \hat{z}_1) & \dots & k_\frac{3}{2}(\hat{z}_n, \hat{z}_n)
\end{pmatrix},\label{eq:kx}
\end{align}
where each entry $k_\frac{3}{2}(\hat{z}_u, \hat{z}_w)$ in Equation~(\ref{eq:kx}) denotes the pair-wise kernel value between $\hat{z}_u$ and $\hat{z}_w$.
\textbf{Finally}, by centering the kernel matrix as described in Equation~(\ref{eq:kx}), we calculate the HSIC value, which helps improve the classification efficacy for $\hat{z}$:
\begin{equation}
\text{HSIC}(\hat{z}, y) = \frac{1}{{(n - 1)}^2} \operatorname{tr} (K_{\hat{z}} H K_y H), \label{eq:hsic}
\end{equation}
where $H = \textbf{I} - \frac{1}{n} \mathbf{1}\mathbf{1}^T
$ is the centering matrix, $\operatorname{tr}(\cdot)$ denotes the trace operation, $y$ is the class label, and $K_y$ denotes the kernel matrix of $y$. It is important to note that the HSIC value computation, achieved through the inner product of kernel matrices, is inherently {\bf dimension-agnostic}. 
Thus we can derive the learning objective combined with Equation~(\ref{eq:hsic}):
\begin{equation}
\mathcal{L}_{H}=\mathcal{L}_{cls} +\text{HSIC}(\hat{z}, y),\label{eq:hsic_value}
\end{equation}
where $\mathcal{L}_{H}$ is the HSIC-based learning objective and $\mathcal{L}_{cls}$ denotes the empirical classification loss~\cite{chi2022infogcn}. 

Up to now, we have arrived at the HSIC-based learning objective in Equation~(\ref{eq:hsic_value}). Besides this, we further engage in the distillation loss to supervise the alignment between the base model and the auxiliary model. Specifically, to supervise the knowledge distillation from the auxiliary model to the base model, we adopt the distillation method of ~\cite{yun2020regularizing}. Our distillation loss is defined as:
\begin{equation}
\begin{aligned}
\mathcal{L}_{D}= \text{KL}( \sigma(f_{\hat{\theta}}(\hat{z}/P)) || \sigma(f_{\tilde{\theta}}(\tilde{z})/P) 
),
\end{aligned}
\end{equation}
where KL is Kullback-Leibler Divergence, $\sigma$ is the softmax activation, $f(\cdot)$ represents the logit of each model, $\hat{\theta}$ and $\tilde{\theta}$ denote the parameters of the base model and the auxiliary model, and $P > 0$ is the distillation temperature.
On the grounds of distillation loss, we incorporate explicit supervision for the auxiliary model using the cross-entropy loss $\mathcal{L}_{CE}$. The entire training objective of our framework is defined as follows:
\begin{equation}
\begin{aligned}
\mathcal{L}_{Total}= \mathcal{L}_{H} + \mathcal{L}_{CE} + \mathcal{L}_{D}.
\end{aligned}
\end{equation}

\subsection{Multi-Stream Ensemble}

As mentioned earlier, the Gaussian kernel width can regulate the influence of distance on joint dependencies. Therefore, we advocate for training the proposed frameworks with multiple kernel widths to finely capture the non-linear dependencies in different action types.  Specifically, a large kernel width is more suitable for modeling actions highlighting distant joint collaborations, such as the ``Kicking" action. In contrast, a small kernel width is adaptable for modeling the dependencies between adjacent joints, such as the localized relationship between the head and neck joints during the ``Sneeze" action. To this end, we propose feeding the frameworks with joint and bone inputs, each of which is trained with both small and large kernel widths. During the inference phase, the softmax scores from these frameworks are averaged to obtain final prediction scores. 
This ensemble method is expected to significantly enhance the comprehensiveness of action recognition 

\section{Experiments}

In this section, we conduct extensive experiments to empirically evaluate the performance of our method on three benchmark action recognition datasets. We aim to answer the following research questions:
\begin{itemize}
\item \textbf{RQ1:} How is our method comparing to state-of-the-art approaches for skeleton-based action recognition?
\item \textbf{RQ2:} How much do different components of the proposed method contribute to its performance?
\item \textbf{RQ3:} What interesting insights and findings can we obtain from the empirical results?
\end{itemize}
Next, we first present the experimental settings, followed by answering the above research questions one by one.


\subsection{Experimental Settings}
\textbf{Datasets.} \quad
We adopt three widely used action recognition benchmark datasets, namely the NTU-RGB+D 60 dataset, the NTU-RGB+D 120 dataset, and the NorthwesternUCLA dataset, to evaluate the proposed method. 

\emph{NTU-RGB+D}~\cite{shahroudy2016ntu} is designed for the skeleton-based action recognition task. It comprises 56,880 video samples, encompassing 60 action classes performed by 40 volunteers. Each video sample represents a single action and contains a maximum of two subjects. The video sample is recorded using three Microsoft Kinect v2 cameras. The dataset provides two sub-benchmarks:
(1) Cross-Subject (X-Sub): data from 20 subjects is used as the training set, while the rest is used as test data.
(2) Cross-View (X-View): it divides the training and test sets
according to different camera views.
\begin{table}[t]
  \centering
  \resizebox{\columnwidth}{!}{
  \begin{tabular}{lccc}
    \hline
    \multirow{2}{*}{\textbf{Methods}} & \multicolumn{2}{c}{\textbf{NTU RGB+D 120}}\\
      & \textbf{X-Sub (\%)} & \textbf{X-Set (\%)}\\
    \hline\hline
    ST-LSTM~\cite{liu2016spatio} & 55.7 & 57.9 \\
    GCA-LSTM~\cite{liu2017skeleton} & 61.2 & 63.3 \\
    ST-GCN~\cite{yan2018spatial} & 70.7 & 73.2 \\
    2s-AGCN~\cite{shi2019two} & 82.9 & 84.9 \\
    DC-GCN+ADG~\cite{cheng2020decoupling} & 86.5 & 88.1 \\
    MS-G3D~\cite{liu2020disentangling} & 86.9 & 88.4 \\
    Dynamic GCN~\cite{ye2020dynamic} & 87.3 & 88.6 \\
    CTR-GCN~\cite{chen2021channel} & 88.9 & 90.6 \\
    MST-GCN~\cite{chen2021multi} & 87.5 & 88.8 \\
    Ta-CNN~\cite{xu2022topology} & 85.4 & 86.8 \\
    EfficientGCN-B4~\cite{song2022constructing} & 88.3 & 89.1 \\ 
    STF~\cite{ke2022towards} & 88.9 & 89.9 \\
    InfoGCN~\cite{chi2022infogcn} & 89.8 & 91.2 \\
    TranSkeleton~\cite{10029908} & 89.4 & 90.5 \\
    FR-Head~\cite{zhou2023learning} & 89.5 & 90.9 \\
    Stream-GCN~\cite{yang2023action} & 89.7 & 91.0 \\
    SkeletonGCL~\cite{huang2023graph} & 89.8 & 91.2 \\
    \hline
    \textbf{Ours}  & \textbf{90.6} & \textbf{91.7} \\
    \hline
    Ours (Joint)  & 86.0 & 87.4 \\
    Ours (Bone)  & 87.6 & 88.9 \\
    Ours (Joint + Bone)  & 89.3 & 90.7 \\
    \textbf{Ours (4 ensemble)}  & \textbf{90.6} & \textbf{91.7} \\
    \hline
    \end{tabular}}
    \caption{Comparisons of the Top-1 accuracy (\%) with the state-of-the-art methods on the NTU RGB+D 120 dataset.}
    \label{table1}
\end{table}

\emph{NTU-RGB+D 120}~\cite{liu2019ntu} is currently the largest 3D skeleton-based action recognition dataset, which expands on the original NTU-RGB+D dataset by adding 60 additional action classes and 57,600 video samples. It consists of a total of 114,480 samples across 120 action classes, performed by 106 volunteers and captured using three Kinect cameras.
The dataset includes two benchmarks:
(1) Cross-Subject (X-Sub120) divides the 106 volunteers into two groups, with 53 subjects assigned to the training set and the remaining to the test set.
(2) Cross-Setup (X-Set120) are divided into the training and test sets based on their IDs. Samples with even IDs are included in the training set, while samples with odd IDs into the test set.

\emph{Northwestern-UCLA}~\cite{wang2014cross} consists of 1,494 video samples divided into 10 classes. The dataset is recorded using three Kinect cameras and features ten actors. Following the evaluation protocol~\cite{wang2014cross}, training samples are obtained from the first two cameras, and the remaining camera is used to capture test samples.

\begin{table}[t]
  \centering
  \resizebox{\linewidth}{!}{
    \begin{tabular}{lcc}
    \hline
    \multirow{2}{*}{\textbf{Methods}} & \multicolumn{2}{c}{\textbf{NTU RGB+D 60}}\\
      & \textbf{X-Sub (\%)} & \textbf{X-View (\%)}\\
    \hline\hline
    IndRNN~\cite{liu2016spatio} & 81.8 & 88.0 \\
    HCN~\cite{liu2017skeleton} & 86.5 & 91.1 \\
    ST-GCN~\cite{yan2018spatial} & 81.5 & 88.3 \\
    2s-AGCN~\cite{shi2019two} & 88.5 & 95.1 \\
    SGN~\cite{zhang2020semantics} & 89.0 & 94.5 \\
    AGC-LSTM~\cite{si2019attention} & 89.2 & 95.0 \\
    DGNN~\cite{shi2019skeleton} & 89.9 & 96.1 \\
    DC-GCN+ADG~\cite{cheng2020decoupling} & 90.8 & 96.6 \\
    Dynamic GCN~\cite{ye2020dynamic} & 91.5 & 96.0 \\
    MS-G3D~\cite{liu2020disentangling} & 91.5 & 96.2 \\
    MST-GCN~\cite{chen2021multi} & 91.5 & 96.6 \\
    CTR-GCN~\cite{chen2021channel} & 92.4 & 96.8 \\
    Ta-CNN~\cite{xu2022topology} & 90.4 & 94.8 \\
    EfficientGCN-B4~\cite{song2022constructing} & 91.7 & 95.7 \\
    STF~\cite{ke2022towards} & 92.5 & 96.9 \\
    InfoGCN~\cite{wang2022skeleton} & 93.0 & 97.1 \\
    TranSkeleton~\cite{10029908} & 92.8 & 97.0 \\
    FR-Head~\cite{zhou2023learning} & 92.8 & 96.8 \\
    Stream-GCN~\cite{yang2023action} & 92.9 & 96.9 \\
    SkeletonGCL~\cite{huang2023graph} & 92.8 & 97.1 \\
    \hline
    \textbf{Ours}  & \textbf{93.7} & \textbf{97.3} \\
    \hline
    \end{tabular}}
\caption{Comparisons of the Top-1 accuracy (\%) with the state-of-the-art methods on the NTU RGB+D 60 dataset.}
  \label{table2}
\end{table}

\noindent\textbf{Implementation details.} \quad
We conducted our experiments using two NVIDIA GeForce GTX 3090 GPUs. Our model was implemented using PyTorch 1.11. To train our framework, we employed stochastic gradient descent (SGD) with 0.9 Nesterov momentum.
For all datasets, we set the total number of training epochs to 120, with the first 5 epochs dedicated to a warm-up strategy to stabilize the training process. The small Gaussian kernel was set to 1 and the large Gaussian kernel was set to 9.
For the NTU-RGB+D and NTU-RGB+D 120 datasets, we set the initial learning rate to 0.1 and applied a decay of 0.1 every 50 epochs. The batch size was set to 128, and the distillation temperature was set to 1.0. 
For the Northwestern-UCLA dataset, we set the initial learning rate to 0.01, with a decay of 0.1 every 50 epochs. The batch size was set to 32, and the distillation temperature was set to 1.0.

\begin{table}[t]
  \centering
  \resizebox{\linewidth}{!}{
\begin{tabular}{lcc}
    \hline
    \multirow{2}{*}{\textbf{Methods}} & \textbf{Northwestern-UCLA} \\
     & \textbf{Top-1 (\%)} \\
    \hline\hline
    Lie Group~\cite{veeriah2015differential} & 74.2 \\
    HBRNN-L~\cite{du2015hierarchical} & 78.5 \\
    Ensemble TS-LSTM~\cite{lee2017ensemble} & 89.2 \\
    SGN~\cite{zhang2020semantics} & 92.5 \\
    AGC-LSTM~\cite{si2019attention} & 93.3 \\
    4s Shift-GCN~\cite{cheng2020skeleton} & 94.6 \\
    InfoGCN~\cite{chi2022infogcn} & 97.0 \\
    CTR-GCN~\cite{chen2021channel} & 96.5 \\
    Ta-CNN~\cite{xu2022topology} & 96.1 \\
    FR-Head~\cite{zhou2023learning} & 96.8 \\
    Stream-GCN~\cite{yang2023action} & 96.8 \\
    SkeletonGCL~\cite{huang2023graph} & 96.8 \\
    \hline
    \textbf{Ours}  & \textbf{97.2} \\
    \hline
    \end{tabular}}
\caption{Comparisons of the Top-1 accuracy (\%) with the state-of-the-art methods on the NW-UCLA dataset.}
    \label{table3}
\end{table}
\subsection{Comparison with Existing Methods (RQ1)}
We first empirically compare the proposed method with the state-of-the-art approaches.
The experimental results are summarized in Tables~\ref{table1}--\ref{table3}.
Table~\ref{table1} and Table~\ref{table2} present results on NTU RGB+D 120 and NTU RGB+D 60 datasets, while Table~\ref{table3} illustrates results on the Northwestern-UCLA dataset.   
The proposed method consistently outperforms state-of-the-art approaches on all three datasets. For instance, on the NTU RGB+D 60 dataset, state-of-the-art method~\cite{huang2023graph} achieves an accuracy of 92.8\%. In comparison, our method achieves a 93.7\% accuracy, representing a significant increase of 0.9\%. Considering NTU RGB+D 60 is an extensively benchmarked dataset, such improvement is quite hard. The significant improvements validate the effectiveness of our work. 
Compared to FR-Head with the same ensemble setup (4-ensemble) on the NTU RGB+D 120 dataset, our model outperforms by a margin of 1.1\% and 0.8\% in cross-subject and cross-set, respectively. These results empirically confirm the superiority of our method in skeleton-based action recognition.
\begin{table}[t]
  \centering
  \resizebox{\linewidth}{!}{
  \begin{tabular}{lccc}
    \hline
  \textbf{Models}  & \textbf{\# Params.} & \textbf{Acc (\%)} \\ 
  \hline\hline
  Base model~\cite{yan2018spatial} &1.3M&83.4 \\
  \hline
  Base model + Auxiliary model~\cite{yan2018spatial} & 2.5M & 84.4 \\
  Base model + Auxiliary model~\cite{shi2019two}  & 5.1M & 84.5 \\
  Base model + Auxiliary model~\cite{cheng2020decoupling}  & 4.6M & 84.6 \\
  Base model + Auxiliary model~\cite{chen2021multi}  & 4.4M & 85.1 \\
  Base model + Auxiliary model~\cite{lee2023hierarchically}  & 2.9M & \textbf{86.0} \\
  \hline
  \end{tabular}}
  \caption{Comparisons of classification accuracies when using different auxiliary models to predict auxiliary motion information.}
  \label{table4}
\end{table}
\begin{table}[t]
\centering
    \resizebox{.9\linewidth}{!}{
    \begin{tabular}{lcc}
    \hline
    \textbf{Methods} & \textbf{Acc (\%)} & \textbf{Declines (\%)} \\
    \hline\hline
    \textbf{$\mathcal{L}_{Total}$} & \textbf{86.0} & - \\
    \hline
    \quad $\text{w/o} \quad \text{HSIC}$  & 85.1 & 0.9 ($\downarrow$) \\
    \quad $\text{w/o} \quad \mathcal{L}_D$  & 85.2 & 0.8 ($\downarrow$) \\
    \quad $\text{w/o}\quad \text{HSIC}, \mathcal{L}_D$ & 84.9 & 1.1 ($\downarrow$) \\
    \hline
  \end{tabular}}
  \caption{Comparison of classification accuracies based on removing the loss term $\text{HSIC}$ (w/o $\text{HSIC}$) or removing the loss term $\mathcal{L}_D$ (w/o $\mathcal{L}_D$) from the total loss function ${\mathcal{L}_{Total}}$.}
  \label{table5}
\end{table}

\subsection{Ablation Study (RQ2)}
To investigate the effect of individual components of the proposed method, we examine the classification accuracy of our model with different configurations.
All experimental ablation studies are conducted on the X-Sub benchmark of the NTU-RGB+D 120 dataset with joint input modality.

\noindent\textbf{The effects of auxiliary motion information.} \quad
To study the effects of auxiliary motion information generated from the auxiliary model, we replace the auxiliary model with the different models while maintaining the base model unchanged. Specifically, we use the models proposed in~\cite{yan2018spatial,shi2019two,cheng2020decoupling,chen2021multi,lee2023hierarchically} as the auxiliary models, which exhibit gradually increasing performance. The base model is as proposed in~\cite{yan2018spatial}. The results are summarized in Table~\ref{table4}. From the third line to the seventh line in the table, there is a clear positive correlation between the performance of the auxiliary model and the classification accuracy. This observation indicates that incorporating motion information into motion features plays a beneficial role in enhancing their discriminability, and that better motion information generated from the auxiliary model leads to higher classification accuracy.

\begin{table}[t]
\centering
\resizebox{.9\linewidth}{!}{
\begin{tabular}{lc}
\hline
{\textbf{Methods}} & {\textbf{Acc (\%)}}\\ 
\hline\hline
Baseline & 83.4  \\ 
\hline
\quad $+\text{HSIC}, \mathcal{L}_{D}$ & 84.7 (1.3$\uparrow$) \\ 
\quad $+ \text{Novel graph}$ & 84.8 (1.4$\uparrow$)  \\ 
\quad $+\text{HSIC}, \mathcal{L}_{D}, \text{Novel graph}$ & {\bf 86.0} (2.6$\uparrow$)  \\ 
\hline
\end{tabular}}
\caption{Comparisons of classification accuracies when applying each component of our method to the baseline.}
\label{table6}
\end{table}
\noindent\textbf{The effects of learning objectives.} \quad
We further validate the effect of learning objectives. 
To confirm that our learning objectives improve test accuracies, we compare the performance of the proposed model trained with different losses by systematically removing each term from the total loss function $\mathcal{L}_{Total}$. The empirical results are presented in Table~\ref{table5}. From the table, we can observe that removing $\text{HSIC}$ (w/o $\text{HSIC}$) and removing $\mathcal{L}_D$ (w/o $\mathcal{L}_D$) results in accuracy drops of 0.9\% and 0.8\%, respectively. Moreover, the performance is significantly decreased by 1.1\% when removing both $\text{HSIC}$ and $\mathcal{L}_D$ (w/o $\text{HSIC}$, $\mathcal{L}_D$). These findings suggest the effectiveness of the proposed learning objectives in enhancing the entire learning process. 

\noindent\textbf{The effects of ensemble streams.} \quad
To study the effect of ensemble streams, we compare the performance of ensembles of frameworks trained with multiple kernel widths. 
In Table~\ref{table1}, we observe an improvement in accuracy as the number of ensemble streams increases. On the cross-subject, accuracies of joint+bone and 4 ensemble (ensemble of frameworks trained with both small and large kernel widths using joint and bone input) increased by 3.3\% and 4.6\%, respectively, compared to the accuracy of using the joint input only. The results imply that multi-stream fusion indeed enhances the comprehensiveness of action recognition, and further increases the recognition performance.

\noindent\textbf{The contributions of each component.}  \quad
We investigate the contribution of each component in the proposed method as illustrated in Table~\ref{table6}.
The baseline was constructed by removing the novel graph from the base model in our method and trained only with $\mathcal{L}_{cls}$. We observe that learning objectives improve baseline accuracy by 1.3\%, and the proposed novel graph leads to a 1.4\% increase in baseline accuracy. When all proposed components are applied together, the baseline accuracy increases by 2.6\%.



\subsection{Analysis (RQ3)}
In this subsection, we present an in-depth visual analysis to study the impact of the HSIC-based learning objective and provide a quantitative analysis of the Gaussian kernel.

\begin{figure}[t]
  \centering
  \includegraphics[width=\linewidth]{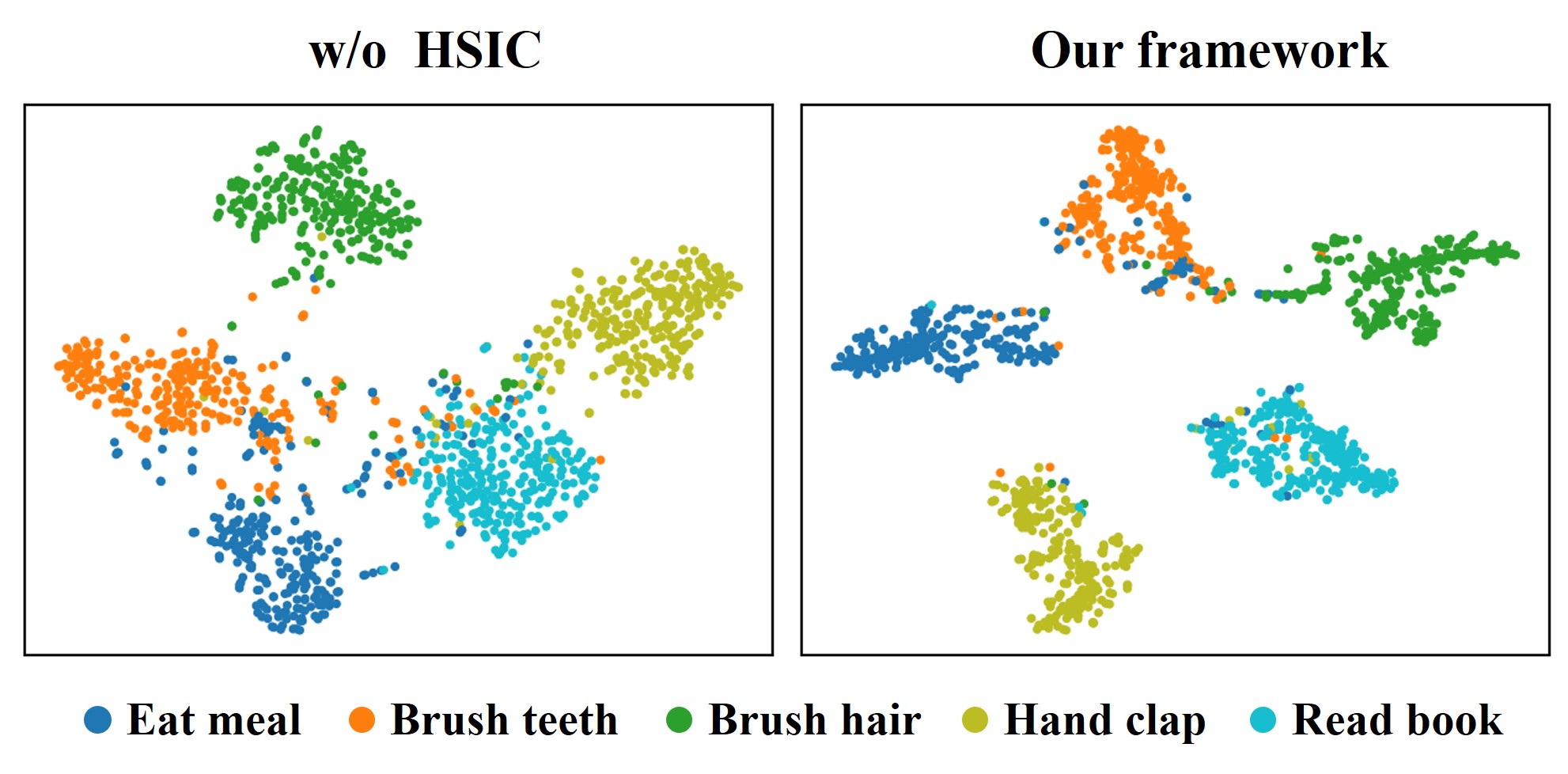}
  \caption{The visualization illustrates feature representations of five randomly selected action classes. We utilize t-SNE for dimension reduction. Each action class is represented by a different color. The five action classes are \emph{Eat meal}, \emph{Brush teeth}, \emph{Brush hair}, \emph{Hand clap}, and \emph{Read book}.}
  \label{fig:vis} 
\end{figure}
\noindent\textbf{Visual analysis of the HSIC-based learning objective.} \quad
To investigate the effect of the HSIC-based learning objective, we perform a visual analysis of the feature representation $\hat{z}$. 
We project these feature representations onto a lower dimensional space with t-SNE and compare them using two different methods: the model trained without or with the HSIC-based learning objective. 

As shown on the left of Fig.~\ref{fig:vis}, when training without the HSIC-based learning objective, the feature representations exhibit overlapping regions, and each action class lacks clear clustering. In contrast, as depicted on the right of Fig.~\ref{fig:vis}, the proposed method effectively separates the five action classes, with the feature representation of each class becoming more distinct. This indicates that utilizing the proposed HSIC-based learning objective facilitates better distinction between different action classes. We observe similar patterns across all other classes but visualize only five randomly selected action classes for simplicity. 

\begin{figure}[t]
\centering
\includegraphics[width=\linewidth]{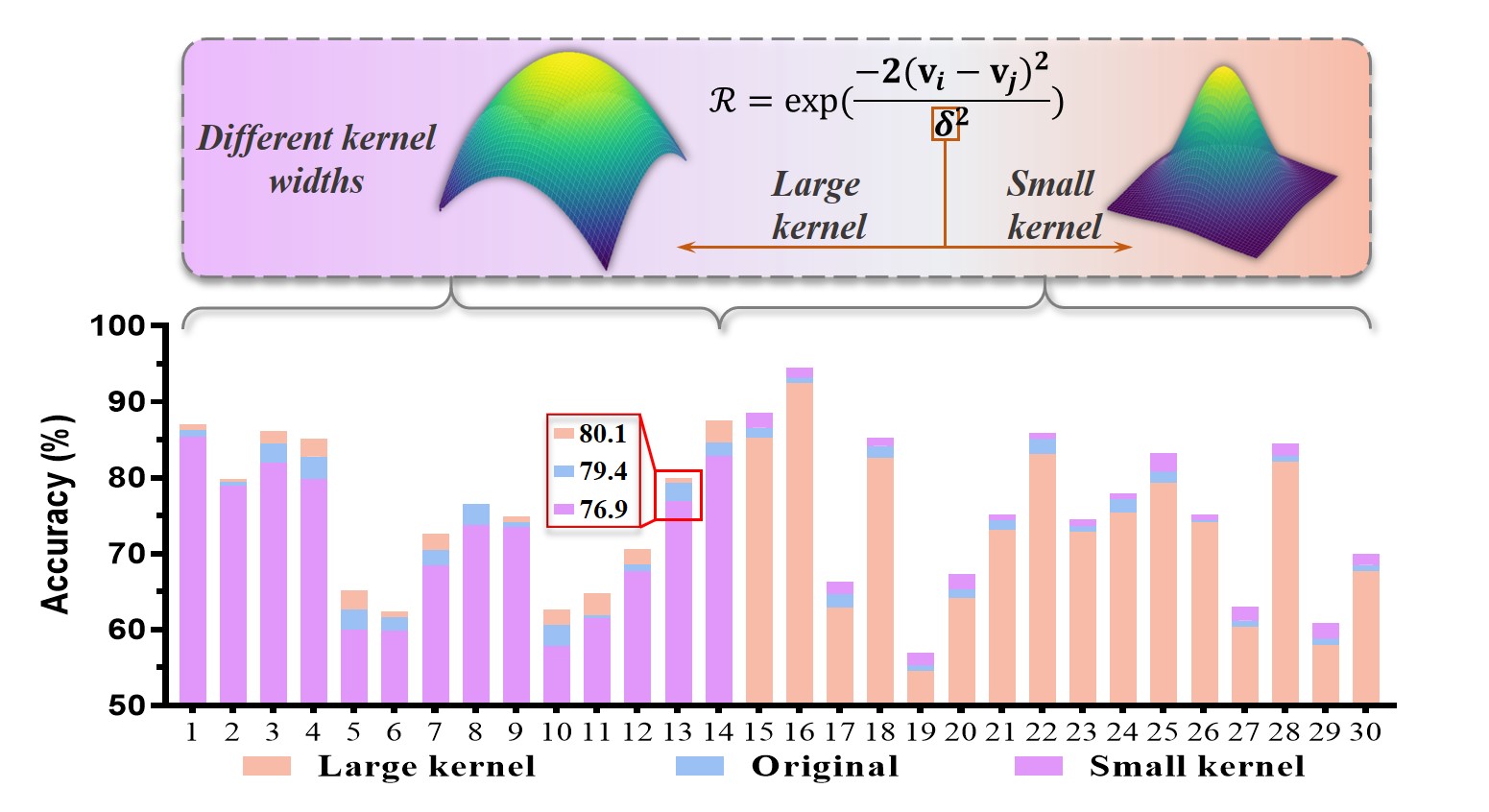}
\caption{The histograms show the quantitative results of applying the three methods to thirty action classes. The quantitative results of each method are depicted with different colors.}
\label{fig:quantitative_results}
\end{figure}

\noindent\textbf{Quantitative analysis of the Gaussian kernel width.} \quad
To study the effectiveness of the kernel width in the Gaussian function, we conduct a quantitative analysis of the recognition results. Specifically, we randomly pick thirty action classes and demonstrate their empirical results in Fig.~\ref{fig:quantitative_results}. We engage three methods in comparison: the framework training with a large kernel width, the framework with the original graph, and the one with a small kernel width.
Taking the ``Staggering" action from the figure as an example, we can observe that the framework using a large kernel yields better results compared to that using a small kernel, leading to a 3.2\% accuracy increase. This finding suggests that using a large kernel is more suitable for modeling the non-linear dependencies of complex actions that involve multiple distant joint coordination, and vice versa.
\section{Limitations and Future Plans}
Despite our method achieving state-of-the-art results on all three datasets, there are still certain aspects that warrant further exploration. 
\emph{Firstly}, expanding our approach to include few-shot learning and unsupervised learning would be fruitful, as it could enhance the applicability of action recognition to real-world scenarios. In our future work, we will concentrate on this aspect. 
\emph{Secondly}, we plan to further study the application of our feature classification method to other tasks~\cite{tang2023duat,tang2024hunting,xu2024polyp,liu2024causality}.

\section{Conclusion}
In this paper, we have proposed a novel dependency refinement method to model the dependencies between human joints. Concretely, we engage in the Gaussian correlation function to capture non-linear dependencies between any pair of joints, mitigating the influence of joint distance. We further propose a framework to distinguish between high-dimensional motion representations and arrive at well-defined learning objectives that ensure the efficacy of the model. Experimental results demonstrate that our method consistently outperforms state-of-the-art methods on three benchmark datasets. 

\section*{Acknowledgments} 
This work was supported in part by the National Natural Science Foundation of China (62276112) and Jilin Province Science and Technology Development Plan Key R\&D Project (20230201088GX).

\bibliography{aaai25}

\end{document}